\documentclass[10pt,twocolumn,letterpaper]{article}
\usepackage[pagenumbers]{cvpr}
\usepackage[T1]{fontenc}
\usepackage{microtype}
\usepackage{array}
\usepackage{amsfonts}
\newcommand{\method}{SemPhys\xspace}

\newcommand{\sg}{\operatorname{sg}}
\newcommand{\E}{\mathbb{E}}
\newcommand{\loss}{\mathcal{L}}


\definecolor{cvprblue}{rgb}{0.21,0.49,0.74}
\usepackage[pagebackref,breaklinks,colorlinks,allcolors=cvprblue]{hyperref}
\ifdefined\XeTeXversion
  \catcode`—=\active
  \def—{---}
  \catcode`–=\active
  \def–{--}
\fi

\title{SemPhys: Semantic-Physical Co-Learning for Humanoid Motion Generation via Differentiable World Models}
\author{%
  \begin{minipage}{\textwidth}
  \centering
  {\large
  Zejian Kang\textsuperscript{1,2}\quad
  Yeqiang Wang\textsuperscript{3,4}\quad
  Zekai Chen\textsuperscript{5}\quad
  Boyang Shen\textsuperscript{6}\\[0.25em]
  Mingyu Ma\textsuperscript{7}\quad
  Bingqian Wu\textsuperscript{2}\quad
  Fan Zhou\textsuperscript{8}\quad
  Jianjie Zhang\textsuperscript{8}\\[0.25em]
  Chentao Shen\textsuperscript{1,2}\quad
  Wentao Yang\textsuperscript{1,2}\quad
  Rui Qian\textsuperscript{9}\quad
  Xiangru Huang\textsuperscript{2,*}\endgraf}
  \vspace{0.65em}
  {\small
  \textsuperscript{1}Zhejiang University\quad
  \textsuperscript{2}Westlake University\\[0.2em]
  \textsuperscript{3}Shanghai Jiao Tong University\quad
  \textsuperscript{4}Shanghai Innovation Institute\\[0.2em]
  \textsuperscript{5}Beijing Institute of Technology\quad
  \textsuperscript{6}Huazhong University of Science and Technology\\[0.2em]
  \textsuperscript{7}Northwest Polytechnical University\quad
  \textsuperscript{8}Xmov\quad
  \textsuperscript{9}Fudan University\\[0.4em]
  \textsuperscript{*}Corresponding author\quad
  \href{mailto:huangxiangru@westlake.edu.cn}{\texttt{huangxiangru@westlake.edu.cn}}\endgraf}
  \end{minipage}%
}
\hypersetup{
  pdftitle={SemPhys: Semantic-Physical Co-Learning for Humanoid Motion Generation via Differentiable World Models},
  pdfauthor={Zejian Kang, Yeqiang Wang, Zekai Chen, Boyang Shen, Mingyu Ma, Bingqian Wu, Fan Zhou, Jianjie Zhang, Chentao Shen, Wentao Yang, Rui Qian, Xiangru Huang}
}

\makeatletter
\renewcommand{\@maketitle}{%
  \begin{center}
  {\LARGE\bfseries\@title\par}
  \vspace{0.8em}
  \@author\par
  \end{center}
}
\makeatother

\begin{document}
\twocolumn[{
\renewcommand\twocolumn[1][]{#1}
\maketitle
\vspace{-0.3em}
\begin{minipage}{\textwidth}
\centering
\captionsetup{hypcap=false}
\includegraphics[width=\textwidth]{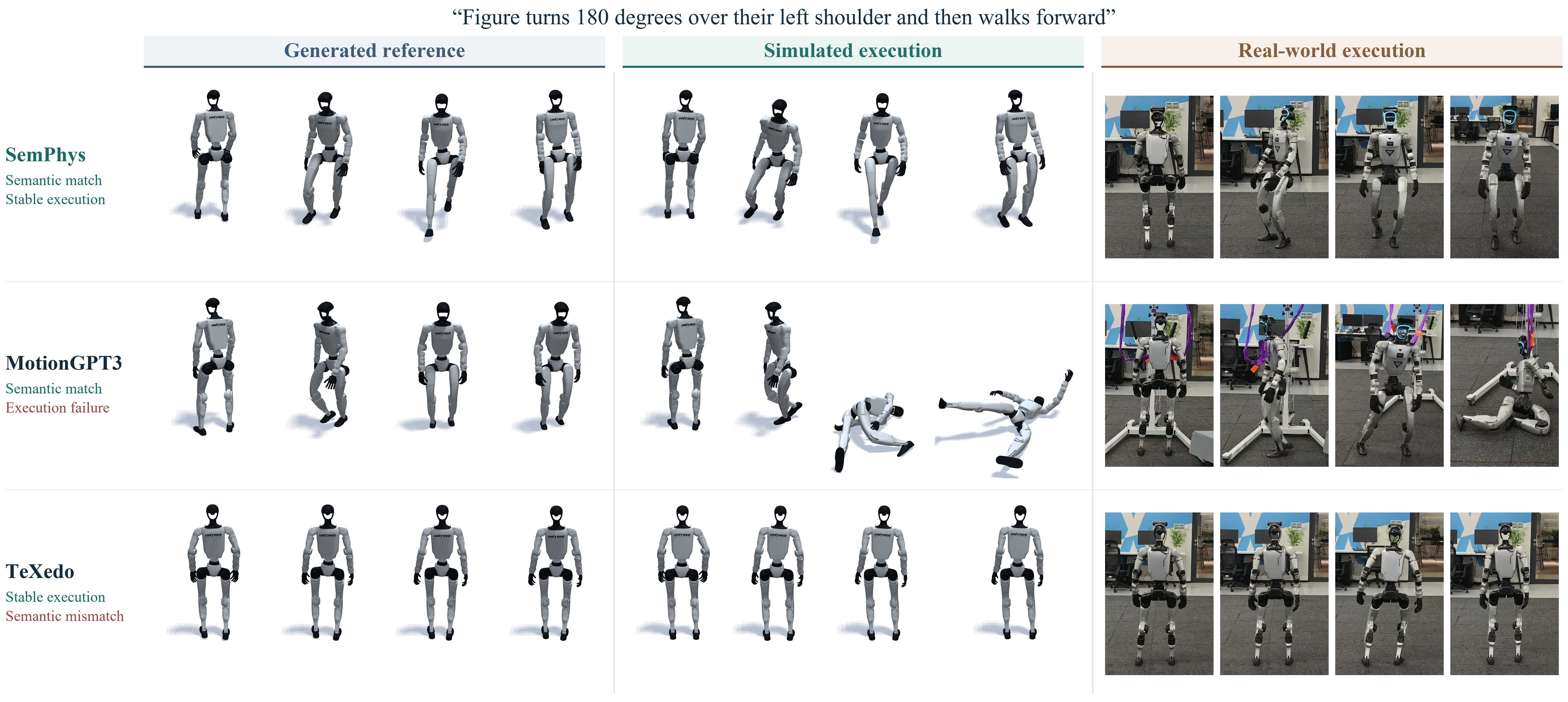}
\captionof{figure}{\textbf{Preserving both semantic fidelity and physical stability during humanoid execution.} Given a composite instruction to turn 180$^{\circ}$ and walk forward, \method{} successfully executes the full sequence in both simulation and the real world. In contrast, existing methods suffer from a trade-off between semantics and stability: MotionGPT3 generates a semantically accurate reference but loses balance during the turn, whereas TeXedo remains stable by staying near its initial pose at the cost of semantic omission. The first two columns compare generated references with simulated trajectories at matched timestamps under a shared frozen tracking controller (initial headings aligned), and the right column shows chronological snapshots from real-world Unitree G1 deployments.}
\label{fig:teaser}
\end{minipage}
\vspace{1em}
}]
\begin{abstract}
In human-robot collaboration, natural language lets people direct a humanoid's whole-body movements without programming joint trajectories. A common approach generates a kinematic reference from an instruction and relies on a tracking controller to execute it. However, as the controller maintains balance under contact constraints, the executed motion can depart from the reference in ways that change the action it conveys. Our central insight is that semantic alignment must be evaluated directly on the complete motion the robot physically executes. Building on this insight, we introduce \textbf{SemPhys}, a framework that trains motion generators using semantic feedback from simulated executions. A differentiable world model and a fixed tracking controller predict a short segment of a recorded execution. SemPhys inserts this prediction at its original position, scores the complete trajectory against the instruction, and backpropagates through the predicted segment. A paired anchoring loss stabilizes adaptation by keeping each generated reference close to its counterpart from the initial generator. On BONES-SEED, SemPhys largely preserves instruction retrieval accuracy from reference to execution and achieves the highest retrieval accuracy on executed motions among the evaluated methods.
\end{abstract}
\section{Introduction}
\label{sec:intro}

Language lets a person direct a humanoid through a sequence such as turning and then walking forward, leaving the robot to coordinate its joints and maintain balance. For the turn-and-walk instruction in Fig.~\ref{fig:teaser}, both parts matter: a robot that starts walking before it finishes the turn travels in a different direction, even if each individual step is stable.

Text-to-motion generators produce reference trajectories~\cite{tevet2023mdm,motiongpt3,momask}, which whole-body controllers track under contact and balance constraints~\cite{luo2023phc,tessler2024maskedmimic,liao2026supermimic}. Execution-aware methods use tracking rewards, candidate ranking, or preferences to improve these references~\cite{serifi2024robotmdm,texedo,zhang2026physmodpo}.

A reference can depict the requested action clearly while leaving the controller to shorten a turn or interrupt a reach to recover balance. Such corrections change the motion that a person watching the robot would associate with the instruction. On BONES-SEED~\cite{bonesseed}, MotionGPT3 retrieves the correct instruction for $74.08\%$ of its references but only $58.18\%$ of their executions. A local correction also depends on the surrounding action: the heading at the end of a turn determines where the next steps lead.

To bridge this gap between kinematic generation and physical execution, \method{} scores the complete execution and backpropagates through a short segment predicted by a differentiable world model and the fixed tracking controller. The predicted poses replace the corresponding portion of a recorded simulator rollout. The prefix tells the scorer how the robot entered the movement, and the suffix shows what followed. For a turn followed by a walk, the scorer can therefore interpret the turn in light of the subsequent steps, while the world model supplies derivatives only through the selected segment. Pairing each update with the initial generator output for the same text and noise stabilizes adaptation.

Extensive evaluations on the BONES-SEED~\cite{bonesseed} and real Unitree G1 hardware demonstrate that \method{} achieves superior executed R@1 of $79.62\%$—surpassing the strongest baseline (MDM) by $19.52$ percentage points—with a minimal sustained failure rate of $0.25$\textperthousand{} on upright tasks, while maintaining robust performance across diverse dynamic perturbations. Real-world deployments confirm that our simulation-trained controller successfully executes complex sequential commands like turns and forward locomotion.

\noindent Our main contributions are summarized as follows:
\begin{itemize}
\item \textit{Semantic supervision on executed actions:} We propose training the generator directly on the semantic match between the instruction and the actual motion produced by the tracking controller, outperforming traditional generation-space scoring.
\item \textit{A local world-model gradient for full actions:} We introduce a short closed-loop prediction embedded within recorded trajectories to backpropagate global semantic loss, preserving global motion context better than static endpoint context filling.
\item \textit{Paired prior anchoring for stable adaptation:} By binding generator updates to initial responses under identical instructions and noise, our method substantially reduces training variance and physical execution failures.
\end{itemize}
\section{Related work}
\label{sec:related}

\paragraph{Execution-aware motion generation.}
RobotMDM predicts tracking rewards~\cite{serifi2024robotmdm}. TeXedo and concurrent physics-in-the-loop selection rank candidate motions~\cite{texedo,memmesheimer2026sample}, while PhysMoDPO learns preferences from tracked executions~\cite{zhang2026physmodpo}. Other methods impose physical constraints during denoising or refinement~\cite{yuan2023physdiff,serifi2024robotmdm,han2025reindiffuse,li2025morph}. For a compound instruction, the semantic consequence of a tracking correction may emerge later in the movement: a shortened turn changes the direction of the walk that follows. \method{} therefore differentiates the instruction score through a local prediction of the tracking loop.

\paragraph{Differentiable dynamics and generative alignment.}
Learned dynamics and short differentiable rollouts provide gradients for policy updates~\cite{heess2015svg,hafner2020dreamer,xu2022shac}. Concurrent SuperMimic backpropagates tracking rewards through a world model to train a whole-body controller~\cite{liao2026supermimic}. We hold the controller fixed and use execution gradients to update the text-conditioned generator. The distinction matters at deployment. The world model supplies derivatives during training, while the tracking policy executes the resulting references on the robot. This connects reward-based generative alignment~\cite{black2024ddpo,wallace2024diffusiondpo,clark2024draft,prabhudesai2024vader,pappa2024modipo} and differentiable motion control~\cite{yao2024moconvq,fu2026icmpg} to sequence-level text--motion matching~\cite{guo2022humanml,petrovich2023tmr}.

\paragraph{Kinematic text-to-motion and whole-body control.}
Text-to-motion models generate continuous poses~\cite{guo2022humanml,petrovich2022temos,tevet2023mdm,zhang2022motiondiffuse}, denoise motion latents~\cite{chen2023mld}, or model discrete motion tokens~\cite{guo2022tm2t,zhang2023t2mgpt,momask,jiang2023motiongpt,motiongpt3}. In physics-based control, AMP learns motion priors and ASE learns reusable skills~\cite{peng2021amp,peng2022ase}. Tracking policies execute reference poses~\cite{peng2018deepmimic,luo2023phc,tessler2024maskedmimic}; humanoid systems extend this interface to teleoperation~\cite{fu2024humanplus,he2024omnih2o} and expressive whole-body tracking on hardware~\cite{ji2024exbody2,he2025asap}.
\begin{figure*}[!t]
\centering
\includegraphics[width=\textwidth]{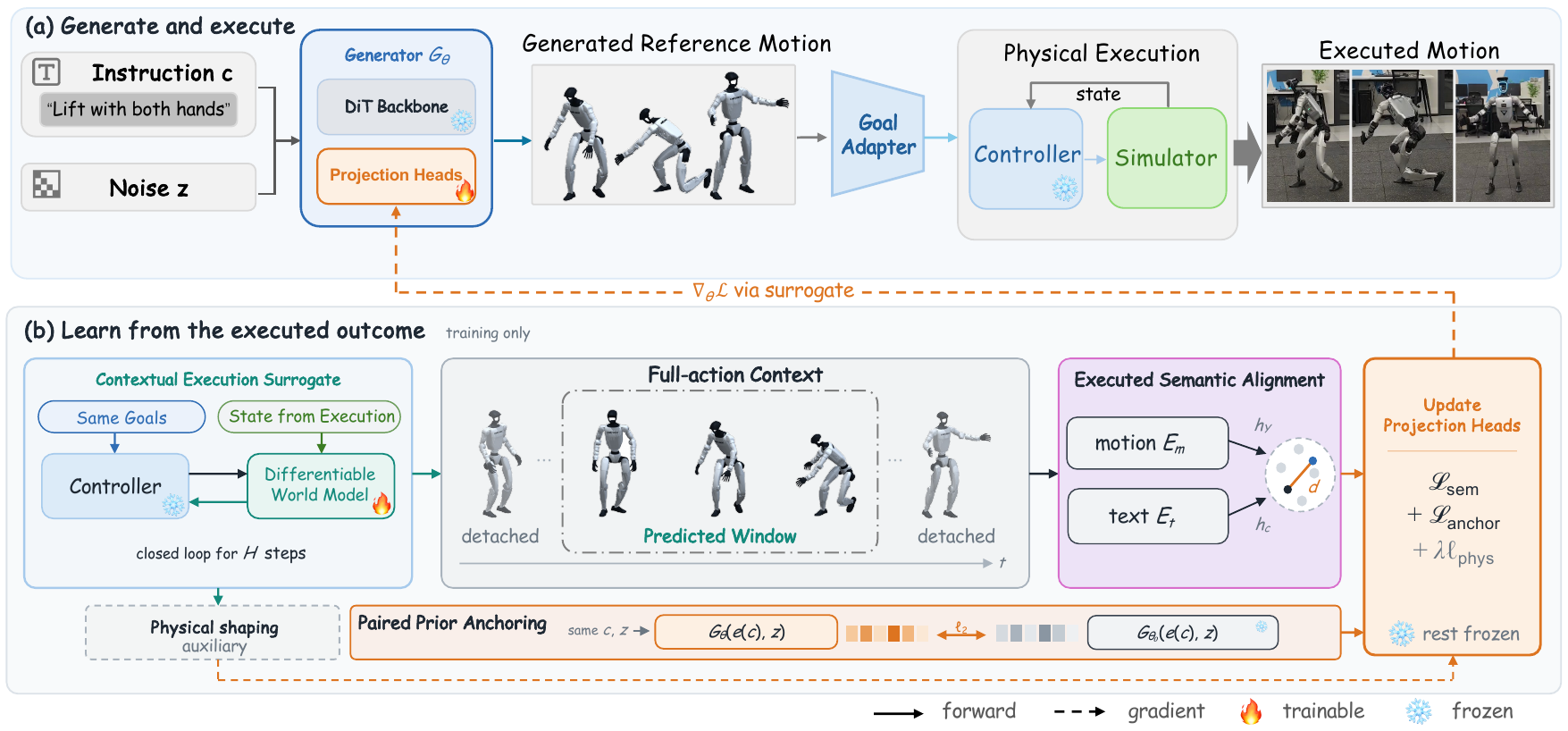}
\caption{\textbf{Overview of \method{} training via execution-space semantic feedback.} \textbf{(a) Generate \& Execute:} A DiT generator synthesizes a reference motion from text and an LLM event plan, tracked by a frozen controller in simulation. \textbf{(b) Surrogate Learning:} A learned differentiable world model and the frozen controller predict a short closed-loop window ($H=32$ at 50\,Hz) from a recorded state. Spliced into the full episode, this window enables a frozen matcher to backpropagate semantic gradients to the generator's projection heads (orange). Prior anchoring and physical shaping regularize updates; snowflakes denote frozen modules. Icons adapted from Smashicons/Flaticon.}
\label{fig:method}
\end{figure*}
\section{SemPhys: semantic and physical learning}
\label{sec:method}

\subsection{Overview and problem formulation}
\label{sec:formulation}
\paragraph{From instruction to executed motion.}
Figure~\ref{fig:method} shows how the generator, tracking controller, and world model interact during training.
Given a natural language instruction $c$ and random noise $z$, the text-conditioned generator produces a reference motion trajectory $X_\theta$ via:
\begin{equation}
z\sim\mathcal N(0,I),\qquad X_\theta=G_\theta(e(c),z).
\label{eq:generator}
\end{equation}
Here, $G_\theta$ incorporates a differentiable pose decoder. The frozen conditioning interface $e$ fuses text embeddings with an LLM-predicted event plan, and a differentiable adapter $g_t$ translates reference poses into tracking targets for the feedback controller $\pi_\phi$. Starting from a neutral state, the controller and physics simulator $\mathcal S$ sequentially generate execution actions and subsequent states:
\begin{align}
u_t&=\pi_\phi(o_t,g_t(X_\theta)),\nonumber\\
s_{t+1}&=\mathcal S(s_t,u_t;\xi),\quad t=0,\ldots,T-1.
\label{eq:execution}
\end{align}
In this formulation, $s_t$, $o_t$, and $u_t$ denote the robot state, history-dependent observation, and control action, respectively, while $\xi$ parameterizes the simulator dynamics. Applying pose extraction $P$ yields the final executed motion trajectory:
\begin{equation}
Y_{\theta,\phi}=\big(P(s_0),\ldots,P(s_{T-1})\big).
\label{eq:executed_motion}
\end{equation}

\paragraph{Learning objective.}
Let $d(c,Y)$ quantify the semantic mismatch between an instruction and a motion trajectory. With a fixed tracking controller $\phi_0$, our optimization objective aims to minimize the expected mismatch over executed trajectories:
\begin{equation}
\min_\theta\ \E_{c,z}\,d(c,Y_{\theta,\phi_0}).
\label{eq:formulation}
\end{equation}
To solve this end-to-end, our framework introduces three core components: \emph{Executed Semantic Alignment} evaluates motion fidelity post-execution; a \emph{Contextual Execution Surrogate} backpropagates gradients through both the controller and a learned local dynamics model; and \emph{Paired Prior Anchoring} regularizes adaptation to prevent generative drift relative to the initial model.

\subsection{Executed semantic alignment}
\label{sec:semantic}
\paragraph{Semantic matching.}
To evaluate semantic alignment, frozen text and motion encoders $E_t, E_m$ project instructions and executed trajectories into a shared latent space~\cite{guo2022humanml,petrovich2023tmr}. A differentiable conversion operator $\mathcal{C}$ maps raw robot poses into the motion encoder's input representation:
\begin{equation}
h_c=E_t(c),\qquad h_Y=E_m(\mathcal C(Y)).
\label{eq:semantic_embeddings}
\end{equation}
While the generator conditions on $e(c)$, the training reward is derived from the distance between $h_c$ and $h_Y$. We define the semantic mismatch $d$ and reward $r_{\mathrm{sem}}$ as:
\begin{equation}
d(c,Y)=\|h_Y-h_c\|_2,\qquad r_{\mathrm{sem}}(c,Y)=-d(c,Y).
\label{eq:reward}
\end{equation}
We refer to this matching distance as MM. Since both encoders yield unit-normalized embeddings, the squared distance simplifies to $d^2=2-2\cos(h_Y,h_c)$. This sequence-level scoring mechanism robustly accommodates temporal variations and pose differences that convey identical semantic intent.

\paragraph{Where supervision acts.}
Evaluating either the executed trajectory $Y_{\theta,\phi_0}$ or its corresponding reference $X_\theta$ under the identical frozen matcher yields two distinct objectives:
\begin{align}
J_{\mathrm{sem}}(\theta)&=\E_{c,z}\,d(c,Y_{\theta,\phi_0}),\label{eq:semantic_target}\\
J_{\mathrm{ref}}(\theta)&=\E_{c,z}\,d(c,X_\theta).\label{eq:reference_target}
\end{align}
As formulated in Eq.~\ref{eq:semantic_target}, tracking corrections and physical constraints are explicitly factored into the loss via the resulting executed motion. Figure~\ref{fig:core_components} visually contrasts these two supervision paradigms under a shared controller.

\subsection{Contextual execution surrogate}
\label{sec:gradient}
Because the semantic interpretation of a localized movement depends heavily on its global context, we preserve the full simulated episode as a background context while predicting only a localized temporal window via learned dynamics.

\paragraph{Local dynamics through feedback control.}
Given a simulator rollout episode $\tau$, we sample a starting index $k$ such that $k+H\le T-1$. Applying the stop-gradient operator $\sg$ to the recorded state and controller history establishes the initial conditions for the prediction window:
\begin{equation}
\hat s_k=\sg(s_k),\qquad
\hat{\mathcal H}_k=\sg(\mathcal H_k).
\label{eq:window_initialization}
\end{equation}
At each step $t=k,\ldots,k+H-1$, we reconstruct observations, compute tracking actions, and predict subsequent states using the learned dynamics model $f_\psi$:
\begin{align}
\hat o_t&=\mathcal O(\hat{\mathcal H}_t,\hat s_t;\xi),\nonumber\\
\hat u_t&=\pi_{\phi_0}(\hat o_t,g_t(X_\theta)),\nonumber\\
\hat s_{t+1}&=f_\psi(\hat s_t,\hat u_t;\xi).
\label{eq:world}
\end{align}
The predicted states and actions are fed back into the controller history. Consequently, gradients propagate smoothly through the inputs of the frozen feedback controller and the calibrated dynamics.

\paragraph{Complete episode, local gradient.}
Embedding the predicted poses at their original temporal indices within the recorded episode yields a hybrid trajectory:
\begin{equation}
\tilde Y=\big[\sg(Y_{0:k}),\ \hat Y_{k+1:k+H},\
\sg(Y_{k+H+1:T-1})\big].
\label{eq:splice}
\end{equation}
Consider the instruction to turn around and then walk forward (Fig.~\ref{fig:teaser}). A window covering the first walking steps shows the gait, while the preceding motion establishes whether the turn came first. Placing the predicted steps back at their original indices preserves this ordering when the semantic loss scores the complete sequence against the instruction:
\begin{equation}
\loss_{\mathrm{sem}}=\E_{c,z,k}\,d(c,\tilde Y).
\label{eq:semantic}
\end{equation}
The recorded prefix and suffix remain fixed during backpropagation. The gradient reaches the generator through $\hat Y_{k+1:k+H}$, so the update reflects how the predicted segment contributes to the meaning of the complete motion. Training episodes span six seconds.

Let $F_{\psi,\phi_0}$ encapsulate the composite mapping of the controller, history updates, and dynamics rollouts, allowing us to express the predicted window as:
\begin{equation}
\hat Y_w=F_{\psi,\phi_0}(g(X_\theta);\sg(\tau),k).
\label{eq:window_mapping}
\end{equation}
For a sampled temporal window, the resulting pathwise gradient is computed via chain rule:
\begin{equation}
g_\theta^{\mathrm{sur}}
=J_G^{\mathsf T}J_g^{\mathsf T}J_F^{\mathsf T}
\nabla_{\hat Y_w}d(c,\tilde Y).
\label{eq:pathwise_gradient}
\end{equation}
Here, $J_G$ differentiates through the final two DDIM denoising steps, and $J_F$ spans the predicted execution window. The initial window states and recorded poses outside the window are treated as constants (detached via stop-gradient). Prior to each generator update cycle, the dynamics model is re-fitted to recent simulator transitions. We validate the resulting surrogate gradient directions by replaying updated references in the full physics simulator (Sec.~\ref{sec:ablation}).

\begin{table*}[!t]
\caption{\textbf{Instruction following on BONES-SEED after execution under a shared G1 controller.} MM and R@$k$ measure text--motion matching distance and retrieval accuracy (\%). FID measures feature-distribution distance. Fail is the sustained-failure rate per thousand upright tasks. Objective columns report mean $\pm$ SD across three generation seeds.}
\label{tab:main}
\centering\small
\setlength{\tabcolsep}{4pt}
\renewcommand{\arraystretch}{1.15}
\begin{tabular*}{\textwidth}{@{\extracolsep{\fill}}lcccccccc@{}}
\toprule
 & \multicolumn{6}{c}{Objective Metrics} & \multicolumn{2}{c}{User Study}\\
\cmidrule(lr){2-7}\cmidrule(lr){8-9}
Method & MM$\downarrow$ & R@1$\uparrow$ & R@2$\uparrow$ & R@3$\uparrow$ & FID$\downarrow$ & Fail (\textperthousand)$\downarrow$ & Motion$\uparrow$ & Semantics$\uparrow$\\
\midrule
TextOp~\cite{xie2026textop} {\scriptsize (CoRL~'26)} & \shortstack[c]{1.279\\[-1pt]{\scriptsize $\pm$ 0.0012}} & \shortstack[c]{17.72\\[-1pt]{\scriptsize $\pm$ 0.19}} & \shortstack[c]{26.73\\[-1pt]{\scriptsize $\pm$ 0.33}} & \shortstack[c]{33.37\\[-1pt]{\scriptsize $\pm$ 0.34}} & \shortstack[c]{0.3404\\[-1pt]{\scriptsize $\pm$ 0.0010}} & \shortstack[c]{9.53\\[-1pt]{\scriptsize $\pm$ 1.15}} & \shortstack[c]{1.68\\[-1pt]{\scriptsize $\pm$ 0.40}} & \shortstack[c]{1.40\\[-1pt]{\scriptsize $\pm$ 0.59}}\\
TeXedo~\cite{texedo} {\scriptsize (arXiv~'26)} & \shortstack[c]{1.250\\[-1pt]{\scriptsize $\pm$ 0.0012}} & \shortstack[c]{22.58\\[-1pt]{\scriptsize $\pm$ 0.13}} & \shortstack[c]{30.27\\[-1pt]{\scriptsize $\pm$ 0.10}} & \shortstack[c]{36.03\\[-1pt]{\scriptsize $\pm$ 0.10}} & \shortstack[c]{0.3037\\[-1pt]{\scriptsize $\pm$ 0.0008}} & \shortstack[c]{\underline{0.46}\\[-1pt]{\scriptsize $\pm$ 0.08}} & \shortstack[c]{1.58\\[-1pt]{\scriptsize $\pm$ 0.43}} & \shortstack[c]{1.40\\[-1pt]{\scriptsize $\pm$ 0.58}}\\
MoMask~\cite{momask} {\scriptsize (CVPR~'24)} & \shortstack[c]{1.180\\[-1pt]{\scriptsize $\pm$ 0.0004}} & \shortstack[c]{36.61\\[-1pt]{\scriptsize $\pm$ 0.11}} & \shortstack[c]{44.46\\[-1pt]{\scriptsize $\pm$ 0.07}} & \shortstack[c]{48.99\\[-1pt]{\scriptsize $\pm$ 0.14}} & \shortstack[c]{0.3521\\[-1pt]{\scriptsize $\pm$ 0.0002}} & \shortstack[c]{1.81\\[-1pt]{\scriptsize $\pm$ 0.75}} & \shortstack[c]{1.84\\[-1pt]{\scriptsize $\pm$ 0.46}} & \shortstack[c]{\underline{1.86}\\[-1pt]{\scriptsize $\pm$ 0.50}}\\
ECHO~\cite{echo} {\scriptsize (arXiv~'26)} & \shortstack[c]{1.177\\[-1pt]{\scriptsize $\pm$ 0.0010}} & \shortstack[c]{26.36\\[-1pt]{\scriptsize $\pm$ 0.23}} & \shortstack[c]{39.36\\[-1pt]{\scriptsize $\pm$ 0.21}} & \shortstack[c]{47.72\\[-1pt]{\scriptsize $\pm$ 0.24}} & \shortstack[c]{0.2242\\[-1pt]{\scriptsize $\pm$ 0.0013}} & \shortstack[c]{2.09\\[-1pt]{\scriptsize $\pm$ 0.16}} & \shortstack[c]{1.64\\[-1pt]{\scriptsize $\pm$ 0.46}} & \shortstack[c]{1.60\\[-1pt]{\scriptsize $\pm$ 0.63}}\\
OMG~\cite{omg} {\scriptsize (RSS Workshop~'26)} & \shortstack[c]{1.060\\[-1pt]{\scriptsize $\pm$ 0.0015}} & \shortstack[c]{50.78\\[-1pt]{\scriptsize $\pm$ 0.21}} & \shortstack[c]{64.67\\[-1pt]{\scriptsize $\pm$ 0.22}} & \shortstack[c]{71.48\\[-1pt]{\scriptsize $\pm$ 0.29}} & \shortstack[c]{0.1441\\[-1pt]{\scriptsize $\pm$ 0.0013}} & \shortstack[c]{2.24\\[-1pt]{\scriptsize $\pm$ 0.65}} & \shortstack[c]{\underline{1.98}\\[-1pt]{\scriptsize $\pm$ 0.48}} & \shortstack[c]{1.82\\[-1pt]{\scriptsize $\pm$ 0.52}}\\
MotionGPT3~\cite{motiongpt3} {\scriptsize (ICLR~'26)} & \shortstack[c]{1.017\\[-1pt]{\scriptsize $\pm$ 0.0005}} & \shortstack[c]{58.35\\[-1pt]{\scriptsize $\pm$ 0.14}} & \shortstack[c]{70.55\\[-1pt]{\scriptsize $\pm$ 0.22}} & \shortstack[c]{75.88\\[-1pt]{\scriptsize $\pm$ 0.29}} & \shortstack[c]{0.1640\\[-1pt]{\scriptsize $\pm$ 0.0003}} & \shortstack[c]{19.65\\[-1pt]{\scriptsize $\pm$ 0.67}} & \shortstack[c]{1.56\\[-1pt]{\scriptsize $\pm$ 0.37}} & \shortstack[c]{1.72\\[-1pt]{\scriptsize $\pm$ 0.50}}\\
MDM~\cite{tevet2023mdm} {\scriptsize (ICLR~'23)} & \shortstack[c]{\underline{1.010}\\[-1pt]{\scriptsize $\pm$ 0.0013}} & \shortstack[c]{\underline{60.10}\\[-1pt]{\scriptsize $\pm$ 0.29}} & \shortstack[c]{\underline{72.24}\\[-1pt]{\scriptsize $\pm$ 0.22}} & \shortstack[c]{\underline{78.15}\\[-1pt]{\scriptsize $\pm$ 0.19}} & \shortstack[c]{\underline{0.1228}\\[-1pt]{\scriptsize $\pm$ 0.0009}} & \shortstack[c]{5.71\\[-1pt]{\scriptsize $\pm$ 0.56}} & \shortstack[c]{1.40\\[-1pt]{\scriptsize $\pm$ 0.31}} & \shortstack[c]{1.82\\[-1pt]{\scriptsize $\pm$ 0.50}}\\
\method{} & \shortstack[c]{\textbf{0.843}\\[-1pt]{\scriptsize $\pm$ 0.0003}} & \shortstack[c]{\textbf{79.62}\\[-1pt]{\scriptsize $\pm$ 0.06}} & \shortstack[c]{\textbf{89.27}\\[-1pt]{\scriptsize $\pm$ 0.07}} & \shortstack[c]{\textbf{92.93}\\[-1pt]{\scriptsize $\pm$ 0.07}} & \shortstack[c]{\textbf{0.0838}\\[-1pt]{\scriptsize $\pm$ 0.0006}} & \shortstack[c]{\textbf{0.25}\\[-1pt]{\scriptsize $\pm$ 0.16}} & \shortstack[c]{\textbf{2.44}\\[-1pt]{\scriptsize $\pm$ 0.37}} & \shortstack[c]{\textbf{2.68}\\[-1pt]{\scriptsize $\pm$ 0.34}}\\
\bottomrule
\end{tabular*}
\end{table*}

\subsection{Paired prior anchoring}
\label{sec:loss}
\paragraph{Corresponding-sample constraint.}
Let $N_\theta$ denote the normalized generator output prior to pose conversion, and maintain a frozen copy of the initial weights $N_{\theta_0}$. Under identical conditioning instructions and noise vectors, we define the anchoring constraint as:
\begin{equation}
\loss_{\mathrm{anchor}}=
\E_{c,z}\|N_\theta(e(c),z)-\sg(N_{\theta_0}(e(c),z))\|_{\mathrm{mean}}^2.
\label{eq:anchor}
\end{equation}
The mean operator aggregates across both temporal frames and feature channels.

A single instruction can admit several valid motions. If the two generators receive different noise samples, the distance between their outputs includes variation that was already present before adaptation. Using the same instruction and noise gives the anchor a specific reference: the motion the initial generator would have produced for this sample. The semantic loss rewards changes that improve the match after execution, while the anchor penalizes how far those changes move the reference from its starting point.

\paragraph{Prior-anchored optimization.}
Combining executed semantic alignment with prior anchoring yields our core training objective:
\begin{equation}
\loss=\loss_{\mathrm{sem}}
      +\lambda_a\loss_{\mathrm{anchor}}.
\label{eq:total}
\end{equation}
The full training loss $\loss_{\mathrm{train}}$ further incorporates auxiliary physical shaping regularizations. Each training iteration collects a fresh rollout episode, fits the dynamics model via $K_\psi$ transition updates, and performs $K_\theta$ generator updates on newly sampled windows according to:
\begin{align}
\tau&\leftarrow\operatorname{Rollout}_{\mathcal S}(G_\theta,\pi_{\phi_0}),\nonumber\\
\psi&\leftarrow\operatorname{Fit}_{K_\psi}(\psi,\sg(\tau)),\nonumber\\
\theta_{\mathcal I}&\leftarrow\operatorname{AdamW}
(\theta_{\mathcal I},\nabla_{\theta_{\mathcal I}}\loss_{\mathrm{train}}).
\label{eq:alternating_updates}
\end{align}
The $\operatorname{Fit}$ operator minimizes the transition prediction loss on recorded simulator transitions. For each sampled window, reference motions are regenerated using the current round's instructions and noise. AdamW updates exclusively target the projection heads indexed by $\mathcal{I}$, while keeping the tracking controller, text-motion matcher, remaining generator parameters, and calibrated dynamics frozen. The appendix specifies the complete hyperparameter schedule.

\paragraph{From training to deployment.}
At deployment, the generator outputs a motion reference from the instruction and event plan, and the tracking controller follows it using robot feedback. Since the world model and semantic matcher are used exclusively during training, the robot can execute the adapted generator's references in real time without online rollouts or score computations.
\section{Experiments}
\label{sec:experiments}

We evaluate \method{} on large-scale humanoid motion benchmarks, severe dynamics perturbations, and real-world Unitree G1 hardware to answer three key questions: (1) Does execution-space supervision improve semantic fidelity and physical stability over current methods? (2) How robust is \method{} against post-tracking semantic drift and dynamics shifts? (3) How do individual components and surrogate gradients contribute to learning efficacy?

\subsection{Experimental setup}
\label{sec:setup}

\paragraph{Dataset.}
We evaluate on the processed BONES-SEED benchmark~\cite{bonesseed}, whose test split comprises 13,302 motion clips, including 13,082 upright tasks used to measure sustained physical failure.

\paragraph{Baselines.}
\label{sec:baselines}
To ensure fair comparison, the identical frozen tracking controller executes every method's generated references on the simulated Unitree G1. ECHO, OMG, TeXedo, and TextOp are evaluated using their official G1 checkpoints~\cite{echo,omg,texedo,xie2026textop}, while MotionGPT3, MDM, and MoMask are trained on the same processed G1 dataset as \method{}~\cite{motiongpt3,tevet2023mdm,momask}. Each generator predicts segment durations or applies its native stopping rule, and test clips are reassembled using source segmentation metadata.

\paragraph{Evaluation metrics.}
The frozen GloVe matcher supplies the training reward and scores all executions. \textbf{MM} measures Euclidean text--motion embedding distance while \textbf{R@$k$} measures top-$k$ instruction retrieval over 20 shared pools of 32 candidates. \textbf{FID} compares executed and ground-truth motion feature distributions. \textbf{Fail} counts sustained failures per thousand upright tasks.

Table~\ref{tab:main} averages objective metrics over three generation seeds per fixed checkpoint. In Fig.~\ref{fig:core_components}, each adapted variant uses the same initial checkpoint and three training seeds. Generation seed 42 is fixed for evaluation, including the single evaluation of the initial checkpoint.

\paragraph{Implementation details.}
We initialize the DiT on paired text--motion data. Adaptation updates its projection heads for three epochs over 84,287 simulation records, with a frozen SuperMimic-style controller~\cite{liao2026supermimic} and separate world-model calibration on simulator transitions. The $H=32$ window spans 0.64\,s at 50\,Hz control and 200\,Hz simulation in MuJoCo.

\subsection{Comparison with current methods}
\label{sec:quantitative}

\paragraph{Full-test performance.}
Table~\ref{tab:main} shows that \method{} retrieves the correct instruction at rank one for $79.62\pm0.06\%$ of executed clips, with an observed mean failure rate of $0.25\pm0.16$\textperthousand{} on upright tasks. MDM, the strongest baseline in executed R@1, reaches $60.10\%$, a system-level gap of 19.52 percentage points (approximate simultaneous 95\% interval $[11.99,27.05]$). \method{} also leads at R@2 and R@3 and achieves the lowest MM (0.843) and FID (0.0838). Giving each distinct instruction equal weight preserves the lead in MM and R@1.
\begin{figure}[!t]
\centering
\includegraphics[width=\columnwidth]{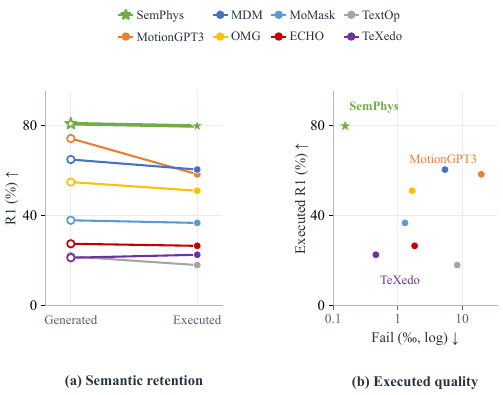}
\caption{\textbf{Semantic fidelity before and after execution.} (a) At generation seed 42, \method{} scores 80.63\% R@1 on its references and 79.59\% on their executions. Lines connect the reference (open) and executed (filled) scores for each method. (b) Executed R@1 and sustained failure (log scale) under a shared frozen controller and nominal dynamics. Retrieval covers the test split, and failure is measured on upright tasks.}
\label{fig:execution_gap}
\end{figure}
\begin{figure}[!t]
\centering
\includegraphics[width=\linewidth]{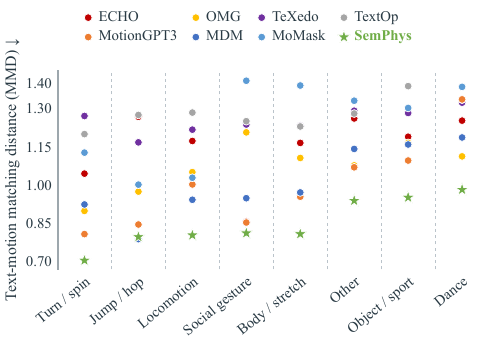}
\caption{\textbf{Executed text--motion matching across eight motion groups.} \method{} (green stars) achieves the lowest matching distance in seven groups. Each group compares all eight methods at generation seed 42. Lower values indicate closer text--motion matching.}
\label{fig:motion_groups}
\end{figure}
\begin{figure}[!t]
\centering
\includegraphics[width=\columnwidth]{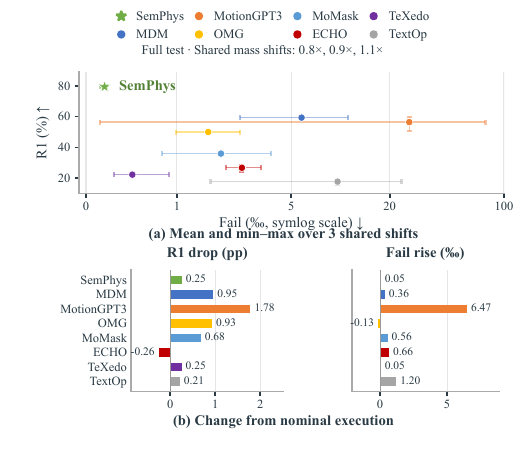}
\caption{\textbf{Retrieval and stability under changes in robot mass.} (a) Means and ranges of executed R@1 and sustained failure across three mass scales ($0.8\times$, $0.9\times$, and $1.1\times$) on the full test set. (b) The mean R@1 drop for \method{} is 0.25 percentage points relative to nominal execution. Negative changes on either axis in (b) indicate improvement.}
\label{fig:robustness}
\end{figure}
\begin{figure*}[!t]
\centering
\includegraphics[width=\textwidth]{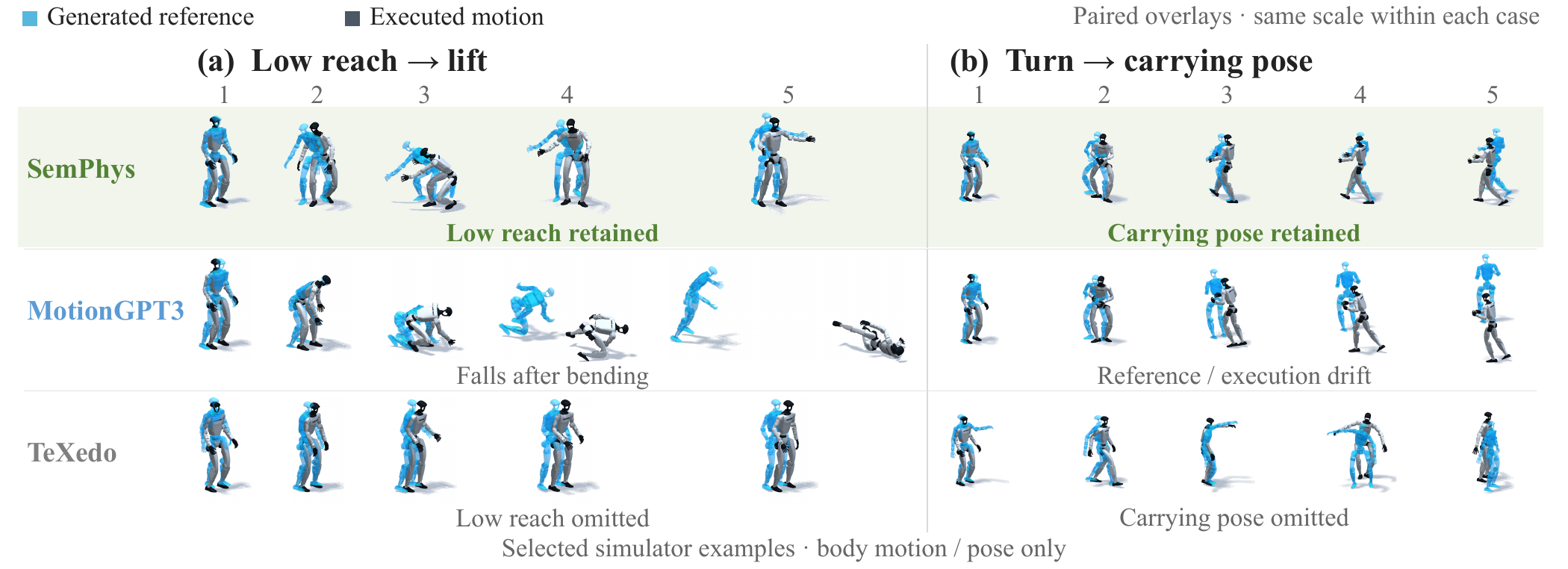}
\caption{\textbf{Preserving the low reach and the carrying pose through execution.} The instructions are (a) ``character picks up big light object with two hands from low height up behind'' and (b) ``character turns back over right shoulder then walks forward while holding medium light object with two hands''. Translucent blue poses show the generated reference and solid robots show its execution at five matched times. The methods share a controller and a display scale within each example. Each trajectory's initial heading is aligned for display. The carrying example illustrates whole-body posture.}
\label{fig:qualitative_execution}
\end{figure*}

\paragraph{Human perceptual evaluation.}
In a blind study of fifty instructions rated by ten respondents, \method{} received the highest average ratings for both motion naturalness ($2.44/3$) and semantic alignment ($2.68/3$), outperforming the runner-up methods OMG ($1.98$) and MoMask ($1.86$) by a clear margin (Table~\ref{tab:main}).

\paragraph{Reducing post-tracking retrieval loss.}
At generation seed 42, we score each reference and its execution under the same controller (Fig.~\ref{fig:execution_gap}). Executed R@1 is $1.04$ percentage points below reference R@1 for \method{} ($80.63\%$ to $79.59\%$). For MotionGPT3, the baseline with the highest reference R@1, the drop is $15.90$ points ($74.08\%$ to $58.18\%$). The paired difference in execution-induced R@1 loss favors \method{} by $14.86$ percentage points (approximate simultaneous $95\%$ interval $[7.07, 22.66]$).

\paragraph{Consistency across diverse motion categories.}
Figure~\ref{fig:motion_groups} groups test clips by text rules applied to their captions and identifiers. Evaluated on the shared subset of clips where all eight methods yield finite matching distances, \method{} achieves the lowest executed MM in seven out of eight categories.

\paragraph{Robustness under dynamics perturbations.}
\method{} exhibits strong resilience to out-of-distribution physics shifts without controller retraining. Across 16 perturbed conditions varying body mass, actuator strength, ground friction, and joint damping, \method{} averages $78.89\%$ executed R@1 and $0.70$\textperthousand{} Fail, outperforming MDM's mean R@1 by $20.43$ percentage points (approximate simultaneous $95\%$ interval $[12.90, 27.97]$). Across the same conditions, TeXedo averages $0.71$\textperthousand{} Fail and $21.57\%$ executed R@1. Figure~\ref{fig:robustness} highlights representative mass perturbations ($0.8\times$, $0.9\times$, and $1.1\times$). The appendix reports all 16 conditions.

\paragraph{Qualitative trajectory comparisons.}
Visual inspection in Fig.~\ref{fig:qualitative_execution} illustrates how \method{} balances expressiveness with physical feasibility. In Fig.~\ref{fig:qualitative_execution}(a), \method{} bends and reaches low with both hands while maintaining balance; MotionGPT3 loses balance and falls after bending, whereas TeXedo remains upright by omitting the low reach. In Fig.~\ref{fig:qualitative_execution}(b), \method{} executes a right-shoulder turn followed by forward locomotion while maintaining a two-handed carrying pose; MotionGPT3 loses tracking mid-turn, while TeXedo drops the sustained carrying posture.

\paragraph{Deployment on Unitree G1.}
\label{sec:hardware}
On the physical G1, the simulation-trained controller follows the turn and then takes forward steps in the new direction (Fig.~\ref{fig:teaser}). We deploy the generator and controller without hardware fine-tuning.

\subsection{Ablations and mechanism analysis}
\label{sec:ablation}

\begin{figure}[!t]
\centering
\includegraphics[width=\linewidth]{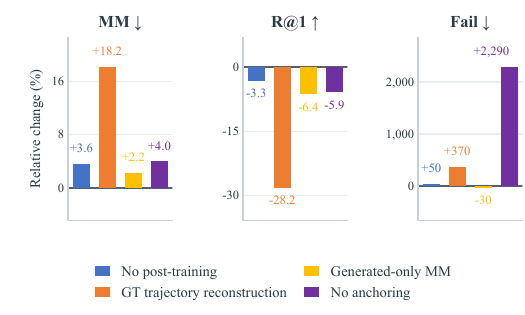}
\caption{\textbf{Supervision on executed motion and paired anchoring improve retrieval.} Each bar shows the relative change $100(m/m_{\mathrm{full}}-1)$ from the full objective. Replacing semantic supervision with GT trajectory reconstruction gives 56.93\% executed R@1 versus 79.30\% for the full objective, a relative change of $-28.2\%$. No post-training denotes the shared first-stage generator.}
\label{fig:core_components}
\end{figure}
\begin{figure}[t]
\centering
\includegraphics[width=\columnwidth]{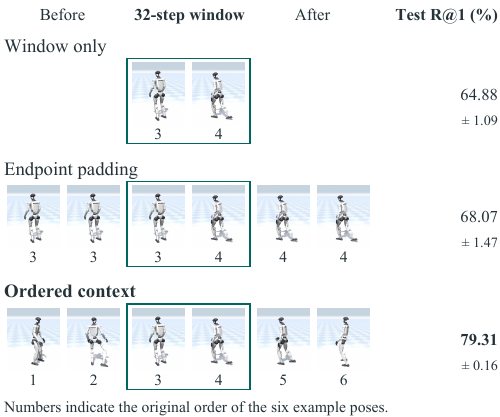}
\caption{\textbf{Motion context during training.} Frames from one G1 simulation illustrate three inputs to the semantic scorer. All rows share the same 32-step differentiable window. Blank regions indicate omitted context. Scores give executed R@1 on the BONES-SEED test split (mean $\pm$ SD across three training seeds).}
\label{fig:context_frames}
\end{figure}
\begin{figure}[!t]
\centering
\includegraphics[width=\columnwidth]{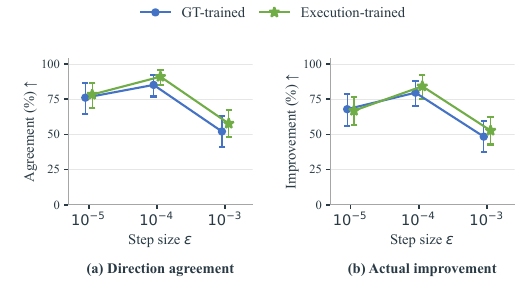}
\caption{\textbf{Testing the surrogate gradient by replaying the updated motion.} (a) Agreement between the predicted and measured directions of change on pairs above replay tolerance. (b) Fraction of finite trials in which the full simulator replay improves matching distance. Error bars show 95\% caption-bootstrap intervals.}
\label{fig:surrogate_validation}
\end{figure}

\paragraph{Impact of execution-space supervision.}
From a shared start of $76.71\%$ executed R@1, supervision on executed motion reaches $79.30\pm0.16\%$ across three training seeds (about $+2.6$ points). The same loss on generated references yields $74.21\pm0.40\%$, a $5.09$-point gap (Fig.~\ref{fig:core_components}). In a separate matched-update control on the same projection heads, execution feedback reaches $79.31\%$ versus $73.31\%$ for continued diffusion supervision (gain $6.00$ points, simultaneous $95\%$ interval $[3.01,9.00]$).

\paragraph{Learning from execution preferences.}
With matched generator updates and nominal simulator transitions, preferences between simulated motions yield $48.54\%$ executed R@1 versus $79.31\%$ for \method{} across three training seeds (paired gain $30.77$ points, simultaneous $95\%$ interval $[20.56,40.97]$). Sustained failures average $0.10$ and $0.23$ per thousand upright tasks, respectively.

\paragraph{Stabilizing adaptation via paired prior anchoring.}
Paired anchoring raises mean executed R@1 from $74.61\%$ to $79.30\%$ and reduces its across-seed standard deviation from $6.35$ to $0.16$ points. Mean Fail falls from $6.09$ to $0.25$\textperthousand{} (Fig.~\ref{fig:core_components}).

\paragraph{Necessity of global temporal context.}
Evaluating the local differentiable window inside its surrounding episode context is essential for sequence-level semantic scoring (Fig.~\ref{fig:context_frames}). Under matched update counts, nominal simulator transitions, and identical 300-frame input lengths, embedding the 32-step predicted window within its original recorded prefix and suffix boosts executed R@1 from $68.07\%$ (with static endpoint padding) to $79.31\%$—an $11.24$-point improvement (simultaneous $95\%$ interval $[4.52, 17.96]$).

Scoring the window alone yields $64.88\%$ R@1. Endpoint padding restores the input length, while the ordered context restores the movement leading into and out of the window.

\begin{table}[!ht]
\caption{\textbf{Sensitivity to gradient window length.} Executed BONES-SEED test scores by window length $H$ in control steps (default 32). Mean $\pm$ SD across three paired training seeds, with generation seed 42.}
\label{tab:window_length}
\centering\small
\setlength{\tabcolsep}{3pt}
\begin{tabular}{@{}rccc@{}}
\toprule
$H$ & R@1 (\%) $\uparrow$ & MM $\downarrow$ & Fail (\textperthousand) $\downarrow$ \\
\midrule
32 & $79.31\pm0.16$ & $0.8453\pm0.0015$ & $0.23\pm0.08$ \\
64 & $\mathbf{80.11}\pm0.32$ & $0.8332\pm0.0021$ & $\mathbf{0.15}\pm0.15$ \\
96 & $79.81\pm0.20$ & $\mathbf{0.8320}\pm0.0002$ & $0.31\pm0.20$ \\
\bottomrule
\end{tabular}
\end{table}

\paragraph{Gradient window length.}
With the full episode scored in every condition, extending the gradient window from 32 to 96 steps reduces MM by 0.0133 (simultaneous 95\% interval for the reduction $[0.0029,0.0264]$). The highest mean R@1 occurs at 64 steps (Table~\ref{tab:window_length}).

\paragraph{Fidelity of surrogate gradients in full simulation.}
We test whether a generator update computed through the local world model changes the score of a fresh, complete simulator execution in the predicted direction. Holding instructions and noise fixed, we step along the surrogate gradient direction ($10^{-4}$ step size) and replay the full trajectory in MuJoCo across 32 validation prompts (Fig.~\ref{fig:surrogate_validation}). Predicted and replayed changes in MM agree in sign for $102$ of $112$ pairs above replay tolerance ($91.1\%$). Furthermore, replayed MM improves in $107$ of $127$ finite trials ($84.3\%$), compared to $48.8\%$ for random perturbations of equal magnitude.
\section{Conclusion}
\label{sec:conclusion}

\method{} trains a text-to-motion generator to preserve the meaning of an instruction through humanoid execution. A differentiable world model lets the generator update a local motion segment using a semantic score for the full trajectory. Paired anchoring keeps each update close to the initial generator's response to the same instruction and noise. On BONES-SEED, the full system achieves $79.62\%$ executed R@1, exceeding MDM by $19.52$ percentage points, with $0.25$ sustained failures per thousand upright tasks.
\end{document}